\documentclass[runningheads]{ola}
\usepackage[utf8]{inputenc}
\usepackage[T1]{fontenc}

\usepackage[noend,ruled]{algorithm2e}

\usepackage{graphicx}
\usepackage[compatibility=false]{caption}
\usepackage{subcaption}
\usepackage{algorithmic}
\usepackage{nicefrac}
\usepackage{color}
\usepackage{cite}
\usepackage{enumerate}
\usepackage{multirow}
\usepackage{booktabs}

\usepackage{amsmath}
\usepackage{amssymb}

\newif\ifusepgf

\usepgftrue

\ifusepgf
\usepackage{pgfplots}
\usepgfplotslibrary{external}
\usepgfplotslibrary{groupplots}
\usetikzlibrary{plotmarks}
\fi

\usepackage{epsfig}
\usepackage{multicol}

\renewcommand{\flat}{{{\mathrm{flat}}}}

\newcommand{\comment}[1]{}

\newcommand{\R}{\mathbb R}

\newcommand{\rev}[1]{#1}

\begin{document}

\pagestyle{headings}

\mainmatter

\title{
Approximation of the objective insensitivity regions using Hierarchic Memetic Strategy
coupled with Covariance Matrix Adaptation Evolutionary Strategy
%Insensitivity detection using Hierarchic Memetic Strategy coupled with CMA-ES
\thanks{The work presented in this paper has been partially supported by
Polish National Science Center grants no.
DEC-2015/17/B/ST6/01867
and by the AGH statutory
research grant no. 11.11.230.124.}
}

\titlerunning{Approximation of the objective insensitivity regions\dots}

\author{Jakub~Sawicki \and Maciej~Smołka \and Marcin~Łoś \and 
  Robert~Schaefer} % \and Piotr~Faliszewski}
\institute{AGH University of Science and Technology\\
Al. Mickiewicza 30, 30-059 Kraków, Poland\\
\email{\{jsawicki,smolka,schaefer\}@agh.edu.pl},
\email{marcin.los.91@gmail.com}}
%\email{\{schaefer\}@agh.edu.pl}}

\authorrunning{J. Sawicki et al.}
%\titlerunning{Two-phase strategy\dots}

\maketitle

\begin{abstract}
\noindent
One of the most
challenging types of ill-posedness in global optimization
is the presence of insensitivity regions in design parameter space,
so the identification of their shape will be crucial, if ill-posedness
is irrecoverable.
Such problems may be solved using
global stochastic search followed by
post-processing of a local sample
and a local objective approximation. % based on B-splines.
We propose a new approach of this type composed of 
Hierarchic Memetic Strategy (HMS)
powered by the Covariance Matrix Adaptation Evolutionary Strategy
(CMA-ES) well-known as an effective, self-adaptable
stochastic optimization algorithm and we leverage the distribution density
knowledge it accumulates to better identify and separate insensitivity regions.
The results of benchmarks prove
that the improved HMS-CMA-ES strategy is effective in both
the total computational cost and
the accuracy of insensitivity region approximation.
The reference data for the tests was obtained by means of
a well-known effective strategy of multimodal stochastic optimization
called the Niching Evolutionary Algorithm 2 (NEA2),
that also uses CMA-ES as a component.

\end{abstract}

\section{Introduction}
\label{sec:Intro}

%\subsection{Motivation for insensitivity regions}
%\label{sec:InsensMotiv}
%\begin{enumerate} [a. ]
%\item 
%{\bf The target Global Optimization Problems (GOPs)}
The Global Optimization Problems (GOPs) are distinguished
among all problems of minimizing real-valued objective function 
$f: \mathcal{D} \rightarrow \mathbb{R}$
over the admissible domain $\mathcal{D}$ embedded in some metric space, 
typically $\mathbb{R}^n, n\geq 1$.
\begin{equation}
\label{eq:GOP}
\mathrm{arg min}_{x \in \mathcal{D}} \{f(x)\}.
\end{equation} 
Roughly saying, GOPs can posses more than one solution, typically due to the
lack of global objective's convexity.
We are focused on ill-conditioned GOPs for which there exist regions in an
admissible domain on which objective takes a value close to the global minimum
and exhibits a very small variability
(insensitivity regions, ``lowlands'', see \cite{Landforms2018} 
for a more formal definition).
%\item
%\end{enumerate}

The ill-conditioned GOPs are frequently encountered when solving Inverse Parametric Problems
(IPPs) tasked to find minimizers to the misfit objective function
with respect to the parameters of an investigated physical, economical, or biological
model.
The misfit is a distance-like function between the observed and simulated
states of a phenomenon under consideration.
There are many important instances of ill-conditioned GOPs studied in the literature.
In particular, the ill-conditioning observed by the irradiative dryer-furnace
design are described in \cite{MIRSEPAHI201777}, while the uncertain
problem of the hydrological rainfall/runoff model calibration was studied in
\cite{Qingyun_1992}. The ambiguity of inverse solutions is also met
by the cancer tissue diagnosis \cite{Majchrzak_2007} and by investigation
of hydrocarbon layers in the magneto-telluric (MT) method \cite{HMSMT2015}.
%%
%% 23.10.2018poniższy a'capite mozna pominac
%%
%The ill-conditioning mentioned above is an irrecoverable feature of the 
%inverse problems, most frequently associated with the restricted observability 
%of a state of phenomena under consideration. It results 
%with the insufficient knowledge, so the unambiguous, smooth data inversion is
%almost impossible. The other reasons are poor formal and computer models of the
%phenomena available. The ill-conditioning is also deepened by measurement
%and computational errors.

The traditional approach of solving ill-conditioned GOPs exhibiting
objective's insensitivity over some regions of admissible domain is the objective
regularization \cite{Beilina2012}. However, this method is strongly restricted to
the problems which are conditionally regular in sense of Tikhonov \cite{Tarantola2005}.
Moreover, the result of such procedure is a single minimizer without
information about the objective behavior in its neighborhood.
An attempt at local objective regularization performed in parallel in
multiple insensitivity regions combined with the advanced multi-deme
memetic search HMS was proposed in \cite{LocalTikhonov2017}.
Another, more general approach
%to solving GOPs with insensitivity regions surrounding global minimizers 
is to deliver an approximation of such sets
\cite{Applanforms2018}.
The strategy followed in this paper consists in generating a random sample of
points located in the insensitive regions, \rev{followed by} assigning these
points toeconnected components of these regions.
After that, a continuous, local objective approximation is generated % separately
for each connected insensitive set component. Finally the level set of such
objective approximation can be taken as an approximated representation of this component.
Of course, the level at which this representation is taken must be sufficiently
close to the minimum objective encountered and this distance is one of the
parameters of the strategy.
Some results obtained in this way were published in
\cite{Applanforms2018}. %,ICCS_2017_irremediable,Evo_2017_Approx}.
%% Poniższy tekst był zduplikowany poniżej w opisie algorytmu
%The local objective approximation was obtained by the Lagrange $1^{\text{st}}$
%order splines on the tetrahedral grid spanned over the local sample's points by
%the Delaunay's algorithm, and next smoothed by projecting on the space of
%$2^{\text{nd}}$ order B-splines spanned over the regular polyhedral grid using
%both $L^2$ and $H^1$ projections.
%The approximation of a Kriging type (see e.g. \cite{Yaochu_2005}) was utilized
%as well.

%\subsection{The main goal of the paper}
%\label{sec:MainGoal}

One of the most important factors that affects efficiency of the
procedure leading to the well approximation of insensitivity sets discussed
above is to obtain high quality sample of points located in each connected
component of the insensitivity set.
As far as localizing such components might be performed by a stochastic search
using low accuracy objective evaluation in the \emph{global phase} of the
strategy, this accuracy should be increased, when the final, local search is
executed, which generates additional significant computational cost in the
\emph{local phase}.
Till now, both phases were engined by the Simple Evolutionary Algorithm (SEA)
with normal mutation and arithmetic crossover \cite{Schaefer2007}.

The main contribution of this paper is to replace SEA in the local phase
by the CMA-ES algorithm, highly appreciated as an effective, self-adaptive
stochastic local search in continuous domains (see e.g. \cite{Hansen2001CMA-ES}).
We will exploit the analytical representation of the sampling measure, 
that is created and adapted in each CMA-ES iteration.
We expect CMA-ES to quickly and cheaply stabilize the density of its
sampling measure, so that the region of its large value covers 
the current connected insensitivity component.
Then, we cover this region by a set of points suitable for the
local objective approximation.
The effective strategy of multimodal stochastic optimization NEA2 \cite{preuss_2010}
applying also CMA-ES as a component will be \rev{used} for obtaining the reference data.
% Nie robimy tego w ten sposób
%While obtaining the optimal approximation grid is nontrivial, we hope to get a
%good quality one in a single, final step of the local phase, by generating
%deterministic, structured mesh, or by generating stochastic points with a
%uniform distribution in the area determined by a CMA-ES density and then reduce
%them by the \emph{multiwinner selection} \cite{IS_2016}.

\rev{
The plan of the paper is as follows.
We introduce the solving strategies in Section~\ref{sec:ProbStat}.
Then, we provide test results in Section~\ref{sec:exp-v2}.
Finally, the conclusions are presented in Section~\ref{sec:Conc}.
}

%\noindent
%+++++++++++++++++++++++++++++++++++++\\
%+++++++++++++++++++++++++++++++++++++

\section{Solving strategies}

\label{sec:ProbStat}

%\subsection{HMS as a global optimizer and an inverse solver}
%\subsection{Hierarchic Memetic Strategy}

\subsection{The global phase strategies}
%Our main tool in solving GOPs and in particular the ill-posed inverse problems
%formulated as GOP is Hierarchic Memetic Strategy (HMS). 
\begin{paragraph}
% \label{sec:HMS}
{\bf Hierarchic Memetic Strategy (HMS)}
is a complex stochastic strategy 
consisting of a multi-deme evolutionary
algorithm and other accuracy-boosting, time-saving and knowledge-extracting
techniques, such as gradient-based local optimization methods, dynamic accuracy
adjustment, sample clustering and additional evolutionary components equipped
with a multiwinner selection operator aimed at the discovery of insensitivity
regions in the objective function landscape (see e.g. \cite{HMSMT2015, %IS_2016,
Applanforms2018} and the references therein).

%In this paper we focus on the evolutionary core of HMS (bearing its own name,
%i.e., Hierarchic Genetic Strategy or HGS).
%For the full description of HGS/HMS we refer the reader to the above-mentioned
%papers together with \cite{HMSMT2015}. %FogaHGS,HGSRN}.
%Here we provide only a concise summary.

%As said above, HGS is a multi-population (multi-deme) strategy.
%Its 
The HMS sub-populations (demes) are organized in a parent--child tree
hierarchy.
The number of hierarchy levels is fixed but the degree of internal nodes is
not.
Each deme is evolved by means of a separate single-population evolutionary
engine such as SEA.
In a single HMS global step (a \emph{metaepoch}) each deme runs a prescribed
number of local steps (genetic epochs).
After each metaepoch, a change in the deme tree structure can happen: some of
the demes that are not located at the maximal level of the tree can produce
child demes through an operation called \emph{sprouting}.
It consists in sampling a set of points around the parent deme's current best
point using a prescribed probability distribution: here we use the normal
distribution.
The sprouting is conditional: we do not allow sprouting new children too close
to other demes at the target HMS tree level.
HMS typically starts with a single parent-less \emph{root} deme.
The maximal-level child-less demes are called \emph{leaves}.
The evolutionary search performed by the root population is the most chaotic
and inaccurate.
The search becomes more and more focused and accurate with the increasing tree
level.
The general idea is that the higher-level populations discover promising
areas in the search domain and those areas are explored thoroughly by the child
populations.
It is then the leaves that find actual solutions.

The hierarchic structure of the HMS search is especially effective
if the computational cost of
objective evaluation strongly decreases with its accuracy,
which is typically the case when solving IPPs.
\end{paragraph}

%\subsection{The evolution of CMA-ES sampling measure} 
%\label{sec:CMA-ES}

\begin{paragraph}
{\bf The CMA-ES stochastic optimization algorithm} performs a stochastic, adaptive
search in a continuous domain $\mathcal{D} \subset \mathbb{R}^n$ for some $n
\geq 1$.
It adapts a multivariate normal distribution iteratively.
Points sampled from this distribution are used to decide how to modify the mean
point contained in $\mathcal{D}$, so that it will be closer to the exact
solution.
The operation of the algorithm is thoroughly described in
\cite{Hansen2001CMA-ES}, we present here only its concise description.

The initial population of the cardinality $\lambda$ is sampled using the normal
distribution with some starting values of a mean $m \in \mathcal{D}$ and a
covariance matrix $\sigma^2 C$, where $\sigma \in \mathbb{R}_+$ is the
parameter scaling the average standard deviation,
$C$ is the $n \times n$ normalized covariance matrix.
Typically, the initial setting of $C$ has unit entries on diagonal and zeros
outside.

In each consecutive iteration $k$, a population $P_k$ is sampled from the
distribution and evaluated.
$P_k$ is then used to modify the representation of the sampling measure in the
next iteration, i.e., new mean $m_{k+1}$, new $\sigma_{k+1}$ and new covariance
matrix $C_{k+1}$ are determined.

Summing up, CMA-ES transforms the density of its sampling measure,
an instance of the multivariate normal distribution parametrized by a tuple
$(m_k, \sigma_k, C_k)$.
At the start of each iteration $k$ it performs 
$\lambda$ - times sampling with return
according to the distribution 
$N(m_k, \sigma_k^2 \, C_k)$
obtaining the population $P_k$.
The analysis of the $P_k$ evaluation (fitness values)
leads to the deterministic computing of the
sampling measure $(m_{k+1}, \sigma_{k+1}, C_{k+1})$
for the next iteration.

%Summing up, CMA-ES transforms the density of its sampling measure,
%an instance of the multivariate normal distribution parametrized by a tuple
%$(m_k, \sigma_k, C_k)$ into another distribution
%$(m_{k+1}, \sigma_{k+1}, C_{k+1})$ in each step.
%%Summing up, CMA-ES transforms the density of its sampling measure, being
%%the instance of normal distribution parametrized by the tuple $(m, \sigma, C)$
%%together with the population $P$ and its cardinality $\lambda$
%%so, the state of CMA-ES at the particular epoch $k > 0$ is characterized 
%%by the larger tuple $(P_k, \lambda_k, (m_k, \sigma_k, C_k))$.
%%Passing from a $k^{\text{th}}$ epoch to the next $(k+1)^{\text{th}}$ one 
%%is performed in two steps: 
%\begin{equation}
%\label{eq:CMA_steps}
%\begin{array}{cc}
%(k, (m_k, \sigma_k, C_k)) & \\
%\downarrow &
%\begin{array}{c} 
%\lambda - \text{times sampling with return}\\
%\text{according to the distribution}\\ 
%N(m_k, \sigma_k^2 \, C_k)\\
%\end{array}
%\\
%P_k & \\
%\vspace{-3pt}\\
%\downarrow & 
%\text{deterministic adaptation step}\\
%\vspace{-3pt}\\
%(k+1, (m_{k+1}, \sigma_{k+1}, C_{k+1})) & \\
%\end{array}
%\end{equation}
%%%
%%%
%%\begin{equation}
%%\label{eq:CMA_steps}
%%\begin{array}{ccccc}
%%(k, (m_k, \sigma_k, C_k)) 
%%& 
%%\stackrel{\longrightarrow}
%%{\begin{array}{c}
%%\lambda - \text{times sampling with return}\\
%%\text{according to the distribution}\\ 
%%N(m_k, \sigma_k^2 \, C_k)
%%\end{array}}
%%&
%%P_k
%%&
%%\stackrel{\longrightarrow}
%%{\text{deterministic adaptation step}}
%%&
%%(k+1, (m_{k+1}, \sigma_{k+1}, C_{k+1}))
%%\end{array}
%%\end{equation}
One of the possible stopping conditions of the CMA-ES algorithm is detecting
stagnation of the density adaptation process, \rev{i.e.} the parameters 
$(m_{k+1}, \sigma_{k+1}, C_{k+1})$ do not differ significantly from
$(m_k, \sigma_k, C_k)$. The other, simpler condition is satisfied if
at least one individual reaches fitness below the assumed
\emph{stopFitness} value.
\end{paragraph}

%\subsection{NEA2}
%\label{sec:NEA2}

%\begin{itemize}
    %\item Ogólnie o tym jak działa: random sampling, klastrowanie i CMA-ES.
    %\item Można też trochę o tym, że prosty mechanizm nastawiony na detekcję
        %możliwie wielu optimów.
    %\item Parametry NBC: b ((4.18) z Preuss2015) i $\phi$ (Fig. 4.16)
    %\item NEA2 ma parametry dla CMA-ES: initial sigma i ew. stopFitness
%\end{itemize}
%\cite{preuss_2015}, \cite{preuss_2010}

\begin{paragraph}
{\bf The strategy NEA2} introduced by Mike Preuss \cite{preuss_2010}
effectively searches for the global minimizers to multimodal GOPs.
It is not responsible for filling basins of attractions as well as the areas of
insensitivity surrounding minimizers by the random sample.
The NEA2 scheme is simple and includes two components: the Nearest
Better Clustering (NBC) algorithm and the stochastic local search CMA-ES.

The NBC is a hierarchic clustering algorithm applied to the random sample of
candidate solutions to the GOP under consideration.
A directed graph is created, which vertices are the individuals.
An edge is created for each individual pointing to its closest better neighbor,
unless no such individual exists.
Next, edges are cut according to two rules.
Firstly, edges which are longer than $\phi$ times the mean length of the edges
in the graph.
Secondly, if an individual has at least 3 incoming edges and its outgoing edge
is longer than $b$ times the median length of the incoming ones.
$\phi$ and $b$ are the parameters of the algorithm.
%Next, the edges which are too long with respect to the mean length of the edge
%in the forest or to the median length of the edges appearing in the direct neighborhood of
%the starting, best individuals are removed.
%The edge reduction procedure is controlled by two parameters $\phi$ and $b$
%being the maximum admissible ratios, respectively.

The following three steps are executed consecutively in the main NEA2 loop:
\begin{itemize}
\item
The random sample is generated with the uniform probability distribution over the whole admissible domain $\mathcal{D}$.
\item
The sample is clustered by NBC in order to obtain clusters included in basins of attraction of separate global minimizers.
\item
The CMA-ES processes are started independently for individuals gathered in each cluster. They are stopped, when the stagnation in evolution is observed.
\end{itemize}

The total strategy is stopped after the global stopping condition is satisfied,
\rev{i.e.} the evaluation budget is exceeded.
%e.g. no new global minimizer was encountered during the assumed number of loop
%execution.
The best individuals in each of the CMA-ES processes \rev{form} the final NEA2 solution.

The setting and tuning of all NEA2 parameters including parameters of both
components NBC and CMA-ES is discussed in \cite[Sec. 4]{preuss_2015}.
\end{paragraph}

\subsection{CMA-ES sampling density in insensitivity regions approximation}
\label{sec:CMA-ES-strategy}

The synthetic representation of the sampling measure delivered by
CMA-ES may be applied for the effective investigation of the sets of
insensitivity and their basins of attraction.
The general behavior of the CMA-ES adaptation is to increase the sampling
measure in the regions of interest, i.e. regions with small objective values.
It might be then conjectured, that a sufficiently low level set of such
density function falls into the insensitivity region associated with a
particular global minimizer.

In order to intensify finding insensitivity regions associated with distant
global minimizers, we will use the HMS strategy, in which CMA-ES search engine
is applied in leaf-demes.
%new
We \rev{also used} NEA2 as the alternative global phase of our complex strategy.
%new
The insensitivity region identification might be performed in the following
steps:

\begin{enumerate}[I.]
\item
We run the HMS strategy for the original GOP (\ref{eq:GOP}), until the global
stopping condition is satisfied, i.e. no new insensitivity regions can be
encountered with a sufficiently large probability.
The CMA-ES leaf-demes will be stopped if they encounter areas of flat fitness,
when they would increase $\sigma$ value.
We will utilize their sampling measure, i.e. covariance matrix, at this point.
\vspace{5pt}

\item
We extract the final mean, $\sigma$ and covariance matrix of each CMA-ES deme,
i.e. $(m,\sigma,C)_i$, where $i \in I$ denotes the deme's index and $I$ is the
set of all leaf-deme indexes.
Then, we join the points generated by all the steps of every CMA-ES deme into
set $\mathcal{Q}$.
Mahalanobis distance measures distance in terms of standard deviations from a
normal distribution to a point.
We consider such distance from distribution $(m,\sigma,C)_i$ to each
point from $\mathcal{Q}$.
The points which are within the distance of 1 (no more than 1-sigma) from
$(m,\sigma,C)_i$ are put in set $\mathcal{Q}_i$, for each $i \in I$.
Steps I, II will be called the HMS \emph{global phase} of the new strategy.
%The points which are within the distance of 1 (no more than 1-sigma) are
%selected forming a family $\{\mathcal{Q}_i\}, i\in I$.
%%% RS
%%% Nie rozumiem powyższego zdania. Pomiedzy czym ta odległośc jest liczona?
%%%% JS
%%%% odległość Mahalanobisa jest liczona pomiędzy dystrybucją a punktem.
%%%% tutaj: między dystrybucją (...)_i oraz punktami z Q
%%% Jak sa wyznaczane poszczególne zbiory Q_i rodziny?
%%%
\vspace{5pt}

%new
\item
In order to obtain the alternative for the HMS global phase 
%family of starting sets for the next phase of the total strategy 
we run the NEA2 for the original GOP (\ref{eq:GOP}).
The final populations of each \rev{CMA-ES run of NEA2} and successfully \rev{stopped}
are accepted now as the
%$\{ \hat{\mathcal{Q}}_i^{NEA2} \}, i \in \hat{I}^{NEA2}$
%alternative, 
family of sets of individuals gathered in the basins of attraction.
They may replace the family $\{ \mathcal{Q}_i \}, i \in I$ in the next steps.
We do not differentiate the notation for data coming from HMS and NEA2
global phases in steps IV - VIII, assuming it as generic, context dependent.
\vspace{5pt}
%new

%\item
%Assuming some density threshold $\delta$, we determine the union of level sets 
%of the sampling measure densities established at each leaf-deme CMA-ES process.
%Let us denote it by $\mathcal{K} \subset \mathcal{D}$. The set $\mathcal{K}$
%is typically not connected, so we denote by $\{\mathcal{K}_i\}, i \in I$ the family
%of its connected components. 
%\vspace{5pt}

%\item
%We will denote by $\mathcal{Q}$ the set of all individuals 
%generated by leaf-deme processes
%during the whole history of evolution that are contained in $\mathcal{K}$.
%The set $\mathcal{Q}$ can be analogously spread into the familly 
%$\{\mathcal{Q}_i\}, i \in I$ so that $Q_i \subset \mathcal{K}_i, \, i \in I$.
%\vspace{5pt}

\item
We try to join the sets
$\mathcal{Q}_i, \mathcal{Q}_j, \, i,j \in I,i \neq j$.
The decision whether both sets of individuals should be joined is obtained on
the base of a proper test, e.g. using hill-valley function,
hollow-ridge version (see \cite{ursem1999multinational}). % and also
%Algorithm 2 in \cite{Applanforms2018}).
\vspace{5pt}

\item
After joining sets of individuals belonging to the common insensitivity region
or the basin of attraction of such region we obtain the reduced family
$\{ \hat{\mathcal{Q}}_i \}, i \in \hat{I} \subset I$.
\vspace{5pt}

\item
We execute for each set of individuals $\hat{\mathcal{Q}}_i, i \in \hat{I}$
%(or $\{ \hat{\mathcal{Q}}_i^{NEA2} \}, i \in \hat{I}^{NEA2}$)
the so called \emph{local phase} \rev{of} the strategy, i.e. perform several steps of the
\emph{MultiWinner Evolutionary Algorithm} (MWEA) (see \cite{Applanforms2018}) %IS_2016})
in order to better cover the insensitivity region with individuals.
In particular we intend to spread the individuals uniformly and reduce gaps.
The resulted set of individuals will be denoted by 
$\widetilde{\mathcal{Q}}_i, i \in \hat{I}$.
%(or by $\widetilde{\mathcal{Q}}_i^{NEA2}, i \in \hat{I}^{NEA2}$).
\vspace{5pt}

\item
The next stage consists in designing the local objective approximation for each
set of individuals $\widetilde{\mathcal{Q}}_i, i \in \hat{I}$ 
%(or $\widetilde{\mathcal{Q}}_i^{NEA2}, i \in \hat{I}^{NEA2}$) 
by the methods
described in \cite{Applanforms2018}. %,ICCS_2017_irremediable,Evo_2017_Approx}.
Primarily, the Lagrange $1^{\text{st}}$ order splines on the tetrahedral grid
spanned over the $\widetilde{\mathcal{Q}}_i$ %(or $\widetilde{\mathcal{Q}}_i^{NEA2}$) 
points by the Delaunay's algorithm are prepared.
Next, this function is mapped on the space of $2^{\text{nd}}$ order B-splines
spanned over a regular polyhedral grid using both $L^2$ and $H^1$ projections.
Both types of projections result in $C^1$ smoothness of the local objective
representation.
The approximation of a Kriging type (see e.g. \cite{Yaochu_2005}) can be
applied as well.
Let us denote by $\widetilde{f}_i$ the local approximation of the objective
associated with the set of individuals $\widetilde{\mathcal{Q}}_i$ for all
$i \in \hat{I}$.
\vspace{5pt}
%(or \vspace{5pt}(or $\widetilde{\mathcal{Q}}_i^{NEA2} \; \forall i \in \hat{I}^{NEA2}$).

\item
Finally, the level set of $\widetilde{f}_i$, taken at a sufficiently low level
with respect to the local minimum encountered, will be taken as the
approximation of the insensitivity set component, associated with the set of
individuals
$\widetilde{\mathcal{Q}}_i$, $i \in \hat{I}$
%(or $\widetilde{\mathcal{Q}}_i^{NEA2}, i \in \hat{I}^{NEA2}$)
\begin{equation}
\label{eq:IS}
\mathcal{IS}_\varepsilon^i \,\, \stackrel{\mathrm{def}}{=} \,\,
\left\{
x \in \mathrm{dom}(\widetilde{f}_i) \subset \mathcal{D}; \;
\widetilde{f}_i(x) \leq 
\mathrm{min}_{y \in \widetilde{\mathcal{Q}}_i}
%(or \widetilde{\mathcal{Q}}_i^{NEA2})}
\{\widetilde{f}_i(y)\} + \varepsilon %IS\_factor 
\right\}, 
\; \forall i \in \hat{I}
\end{equation}
where $\varepsilon$ is the tolerance parameter, that determines the maximum
variability of the local objective approximation $\widetilde{f}_i$ which
shouldn't be handled as an essential one.
\end{enumerate}

\section{Experimental verification}
\label{sec:exp-v2}

We will show three benchmark cases.
The first will use a fitness function, which contains 4 non-convex regions of
insensitivity.
We will show how particular stages of the algorithm up to the local
approximation (VII) work on this example.
The second will use a fitness function with similar features, but with more
regions of insensitivity --- we present a comparison based on metrics for this
case.
The third one uses a 4D Rastrigin function to show how higher-dimensional cases
are handled by the strategy.
In all these benchmarks both insensitivity regions and their approximations are constructed
as sublevel sets with cutoff 0.1, i.e. subsets of the domain where the objective function
(or its approximation) assumes values less than 0.1.

%\begin{figure}
    %\centering
    %\begin{tikzpicture}
        %\begin{axis}[
                %width=0.5\textwidth,
                %colormap/blackwhite, colorbar,
                %shader=interp,
                %view={0}{90},
                %xlabel=$x$,
                %ylabel=$y$,
                %%ytick={-2,-1,0,1,2},
                %%xtick={-2,-1,0,1,2},
                %xmin=-20,xmax=20,
                %ymin=-20,ymax=20]
            %\addplot3[surf] file {data/benchmark2/objective.values};
        %\end{axis}
    %\end{tikzpicture}
    %\caption{
        %The benchmark2.
    %}
    %\label{fig:fun2}
%\end{figure}

\subsection{First case - 4 regions of insensitivity}
\label{sec:case-I}

\begin{figure}
    \centering
    \begin{tikzpicture}
        \begin{groupplot}[group style={
                group size=2 by 1,
                horizontal sep=1.5cm,
                vertical sep=2cm},
            width=7cm,
            colormap/blackwhite,
            shader=interp,
            view={0}{90},
            xlabel=$x$,
            ylabel=$y$,
            view={0}{90}]
            \nextgroupplot[
                title={First case},
                xmin=0,xmax=6,
                ymin=0,ymax=6
            ]
            \addplot3[surf] file {data/benchmark3-case-I/objective.values};
            \nextgroupplot[
                title={Second case},
                xmin=0,xmax=20,
                ymin=0,ymax=20,
                colorbar
            ]
            \addplot3[surf] file {data/benchmark4-case-II/objective.values};

        \end{groupplot}
    \end{tikzpicture}
    \caption{
        The plots of the first and the second benchmark functions in their domains.
    }
    \label{fig:fun3-4}
\end{figure}
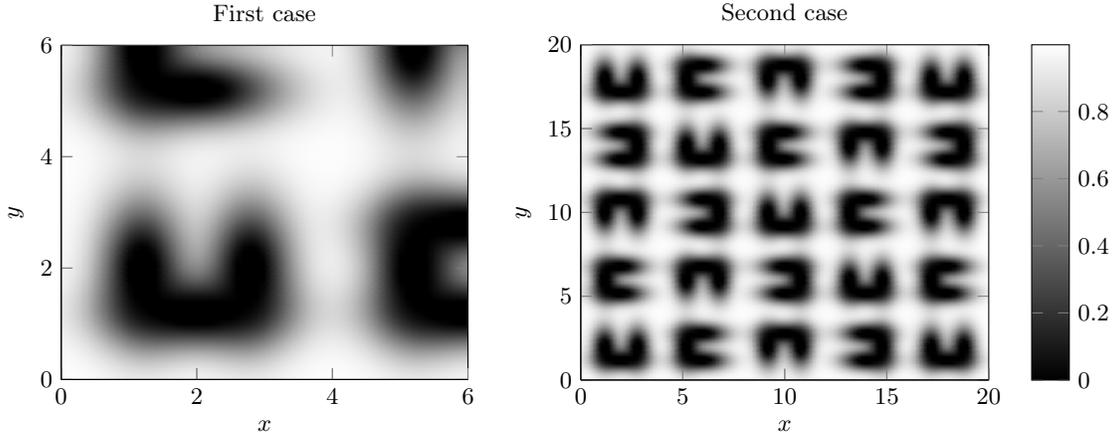

The fitness function is shown in Figure~\ref{fig:fun3-4} and is defined as follows.
We build the benchmark problem by using inverted Gaussian functions defined
like that:
\begin{equation}
    \label{eq:g}
    g^r_{x^0}(x) =
        1-\exp\left(
            -\ln(2)
            (x-x^0)^T
            S
            (x-x^0)
        \right),
\end{equation}
where
$r\in\R_+^N$
and
$S$ is a diagonal matrix with
$S_{i,i}=\nicefrac{1}{r_i^2}$, $S_{i,j}=0, i \neq j$.
We multiply it thrice to obtain a C-shaped valley:
\begin{equation}
    \label{eq:c}
    c(x) = g^{[0.5,1]}_{[-0.8,0]} \cdot g^{[1,0.5]}_{[0,-0.8]} \cdot g^{[0.5,1]}_{[0.8,0]}
\end{equation}
We then define a rotated version of $c$:
\begin{equation}
    \label{eq:c-rot}
    c^\phi(x) = c((x_1\cos\phi-x_2\sin\phi,x_1\sin\phi+x_2\cos\phi))
\end{equation}

And then
\begin{equation}
    \label{eq:h1}
    h_1(x) = \prod_{i=0}^1 \prod_{j=0}^1 c^\theta(x-(2+4i,2+4j))\text{, where }\theta=\frac{\pi}{2}(i+j \mod 4)\text{ ,}
\end{equation}
\begin{equation}
    \label{eq:flat}
    \flat^T_f\colon x \to max\left(\frac{f(x)-T}{1-T},0\right)\text{ ,  }T\in\R\text{ ,}
\end{equation}
and we build the first benchmark function
\begin{equation}
    \label{eq:f1}
    f_1\colon [0,6]^2 \ni (x_1, x_2) \to \flat^{0.1}_{h_1} \in [0,1)\text{ .}
\end{equation}

We compare two algorithms: NEA2 and HMS-CMA-ES, each with attached local phase.
The global phase was stopped when \rev{it} exceeded the budget of 500 evaluations.

HMS has two levels and a metaepoch length of 3.
The root level was SEA with population of 40 individuals,
crossover probability 0.1, normal
mutation with probability 0.5 and standard deviation 2.
Sprout was allowed if no other deme or other sprout seed were found within
Euclidean distance of 1 and its fitness was below 0.5.
The second level was CMA-ES with initial sigma 0.5 which was stopped in case
the average objective didn't change more than 0.01 in 3 epochs.
Moreover, CMA-ES demes were also terminated if they began to increase the sigma
value.
CMA-ES instances used in NEA2 also had initial sigma 0.5.

The hollow-ridge cluster merging method used 3 intermediate points.
The reduced clusters' sizes were used as the MWEA population size if they were
inside the range [10,100].
Otherwise, the population was either randomly sampled to reduce its size to 100
or the missing individuals were generated during the first epoch of MWEA.

The MWEA demes formed from the clusters were stopped after 3 epochs.
The mutation standard deviation was set to $\nicefrac{1}{2}$ of the initial
population's diameter.

\begin{figure}
    \centering
    \begin{tikzpicture}
        \begin{groupplot}[group style={
                group size=2 by 3,
                horizontal sep=1.5cm,
                vertical sep=2cm},
            width=6cm,
            xlabel={x},
            ylabel={y},
            view={0}{90},
            xmin=0,xmax=6,
            ymin=0,ymax=6]

            \nextgroupplot[title={NEA2 selected clusters}]
            \addplot[only marks,mark=x]        table [x=x, y=y] {data/benchmark3-case-I/log-2/nea2-cluster-0};
            \addplot[only marks,mark=+]        table [x=x, y=y] {data/benchmark3-case-I/log-2/nea2-cluster-1};
            \addplot[only marks,mark=o]        table [x=x, y=y] {data/benchmark3-case-I/log-2/nea2-cluster-2};
            \addplot[only marks,mark=asterisk] table [x=x, y=y] {data/benchmark3-case-I/log-2/nea2-cluster-3};
            \addplot[only marks,mark=star]     table [x=x, y=y] {data/benchmark3-case-I/log-2/nea2-cluster-4};
            \addplot[only marks,mark=-]        table [x=x, y=y] {data/benchmark3-case-I/log-2/nea2-cluster-5};
            \addplot[only marks,mark=|]        table [x=x, y=y] {data/benchmark3-case-I/log-2/nea2-cluster-7};
            \addplot[no markers]               table [x=x, y=y] {data/benchmark3-case-I/objective.values.contour};

            \nextgroupplot[title={HMS-CMA-ES selected clusters}, ylabel={}]
            \addplot[only marks,mark=x]        table [x=x, y=y] {data/benchmark3-case-I/log-2/hms-cluster-0};
            \addplot[only marks,mark=asterisk]        table [x=x, y=y] {data/benchmark3-case-I/log-2/hms-cluster-2};
            \addplot[only marks,mark=+]        table [x=x, y=y] {data/benchmark3-case-I/log-2/hms-cluster-4};
            \addplot[only marks,mark=o]        table [x=x, y=y] {data/benchmark3-case-I/log-2/hms-cluster-6};
            \addplot[only marks,mark=|]        table [x=x, y=y] {data/benchmark3-case-I/log-2/hms-cluster-8};
            \addplot[no markers]               table [x=x, y=y] {data/benchmark3-case-I/objective.values.contour};

            \nextgroupplot[title={NEA2 reduced clusters}]
            \addplot[only marks,mark=x]        table [x=x, y=y] {data/benchmark3-case-I/log-2/nea2-r-cluster-0};
            \addplot[only marks,mark=+]        table [x=x, y=y] {data/benchmark3-case-I/log-2/nea2-r-cluster-1};
            \addplot[only marks,mark=o]        table [x=x, y=y] {data/benchmark3-case-I/log-2/nea2-r-cluster-2};
            \addplot[no markers]               table [x=x, y=y] {data/benchmark3-case-I/objective.values.contour};

            \nextgroupplot[title={HMS-CMA-ES reduced clusters}, ylabel={}]
            \addplot[only marks,mark=x]        table [x=x, y=y] {data/benchmark3-case-I/log-2/hms-r-cluster-0};
            \addplot[only marks,mark=+]        table [x=x, y=y] {data/benchmark3-case-I/log-2/hms-r-cluster-1};
            \addplot[only marks,mark=o]        table [x=x, y=y] {data/benchmark3-case-I/log-2/hms-r-cluster-2};
            \addplot[only marks,mark=asterisk] table [x=x, y=y] {data/benchmark3-case-I/log-2/hms-r-cluster-3};
            \addplot[no markers]               table [x=x, y=y] {data/benchmark3-case-I/objective.values.contour};

            \nextgroupplot[title={NEA2 MWEA demes}]
            \addplot[only marks,mark=x]        table [x=x, y=y] {data/benchmark3-case-I/log-2/nea2-lba-0};
            \addplot[only marks,mark=+]        table [x=x, y=y] {data/benchmark3-case-I/log-2/nea2-lba-1};
            \addplot[only marks,mark=o]        table [x=x, y=y] {data/benchmark3-case-I/log-2/nea2-lba-2};
            \addplot[no markers]               table [x=x, y=y] {data/benchmark3-case-I/objective.values.contour};

            \nextgroupplot[title={HMS-CMA-ES MWEA demes}, ylabel={}]
            \addplot[only marks,mark=x]        table [x=x, y=y] {data/benchmark3-case-I/log-2/hms-lba-0};
            \addplot[only marks,mark=+]        table [x=x, y=y] {data/benchmark3-case-I/log-2/hms-lba-1};
            \addplot[only marks,mark=o]        table [x=x, y=y] {data/benchmark3-case-I/log-2/hms-lba-2};
            \addplot[only marks,mark=asterisk] table [x=x, y=y] {data/benchmark3-case-I/log-2/hms-lba-3};
            \addplot[no markers]               table [x=x, y=y] {data/benchmark3-case-I/objective.values.contour};
        \end{groupplot}
    \end{tikzpicture}
    \caption{
        The strategy steps visualised for NEA2 and HMS-CMA-ES global phase.
        Steps are presented in consecutive rows: clusters after the global
        phase (I-II/III), clusters after reduction (IV-V) and points obtained by MWEA (VI).
        The clusters from the first row are not exhaustive to keep the plots
        readable.
        Each cluster is shown with a different mark type and the solid lines are 0.1
        isolines of the fitness function.
    }
    \label{fig:case-I-results}
\end{figure}
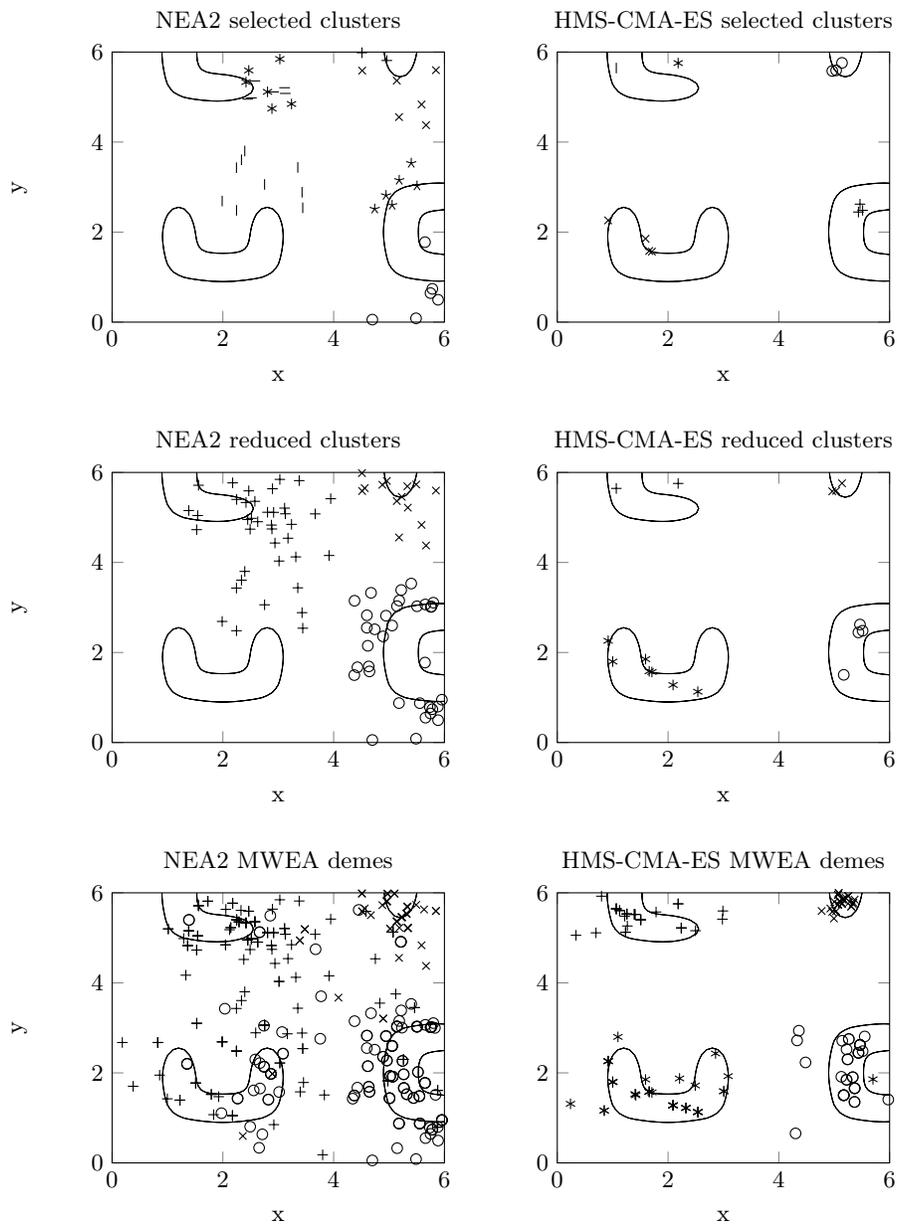

The results of the run are shown in the Figure~\ref{fig:case-I-results}.
The first row shows the clusters which are produced by the global phase, from
left, NEA2 and HMS-CMA-ES.
The second row depicts clusters after they are reduced by the hill-valley
method and the last one shows the points accumulated during the MWEA operation.

In this particular case, HMS-CMA-ES manages to obtain better localized samples,
which separate basins of attraction better than NEA2.
NEA2 generated clusters with more individuals, one of which spanned more than
one basin of attraction.
The resulting MWEA demes' in NEA2 variant as well separate the basins of
attraction worse.

%The level sets for the points from MWEA demes are shown in
%Figure~\ref{fig:case-I-results-approx}.
%Not all the demes for HMS-CMA-ES variant are presented as not all produced sensible results.
%NEA2 level sets are much more accurate than the ones for HMS-CMA-ES.
%However, the NEA2 deme covered the entire domain, instead of being more selective.
%Such behavior suggests, that it will behave similarly for a case with more
%basins of attraction.

%\begin{figure}
%    \centering
%    \begin{subfigure}[b]{0.49\textwidth}
%        \includegraphics[width=\textwidth]{data/benchmark3-case-I/approx-v2/nea2}
%        \caption{Resulting contours for NEA2: cluster 1}
%    \end{subfigure}
%    \begin{subfigure}[b]{0.49\textwidth}
%        \includegraphics[width=\textwidth]{data/benchmark3-case-I/approx-v2/cmaes}
%        \caption{Resulting contours for HMS-CMA-ES: cluster 3}
%    \end{subfigure}
%    \caption{
%        Each graph presents contours of the insensitivity region generated by
%        approximation methods compared to the exact contour.
%        $L^2$, $H^1$ and Kriging methods are shown.
%    }
%    \label{fig:case-I-results-approx}
%\end{figure}

\subsection{Second case - 25 regions of insensitivity}
\label{sec:case-II}

The second benchmark function is shown in Figure~\ref{fig:fun3-4}.
We build it using \eqref{eq:c-rot} and \eqref{eq:flat}:
\begin{equation}
    \label{eq:h2}
    h_2(x) = \prod_{i=0}^4 \prod_{j=0}^4 c^\theta(x-(2+4i,2+4j))\text{, where }\theta=\frac{\pi}{2}(i+j \mod 4)\text{ ,}
\end{equation}
\begin{equation}
    \label{eq:f2}
    f_2\colon[0,20]^2\in x\to\flat^{0.1}_{h_2} \in [0,1)\text{ .}
\end{equation}

The configurations of the algorithms stay the same, apart from the flexible
budget in this test.
The budget will vary from 2000 to 10000 evaluations.
For each budget, each variant is run 10 times to collect statistics of metrics.

We assess the algorithm variant performance by analyzing several factors.
The effect of running a global phase (I-III) is a set of clusters $\mathcal{Q}_i$.
After reducing them (IV-V) to $\hat{\mathcal{Q}}_i$ their number should be of the order of the
number of insensitive regions --- then it is possible to have a single cluster
cover a single region.

After running MWEA (VI) we obtain sets $\tilde{\mathcal{Q}}_i$.
We asses their quality using the metric of
%There are three ways to assess the quality of them in the context of the approximation phase.
%
%Firstly:
the \emph{ratio of covered regions}.
A region is said to be covered by a cluster $\widetilde{\mathcal{Q}}_i$ if every ellipsoid of
the region contains at least a single point from $\widetilde{\mathcal{Q}}_i$.
An ellipsoid has axes of length 1 and 0.5.
The ratio is calculated as a fraction of the covered regions to the total
number of regions, i.e. 25.

%Secondly: the \emph{multiply touched regions ratio}.
%A region is said to be touched by a deme, if a point from the deme covers at
%least one ellipsoid of the region.
%So, if there are two demes which touch a region, then they probably should have
%been reduced to a single cluster in the clustering phase.
%However, this may also be an effect of running MWEA in the local phase.
%This metric is also calculated as a fraction to the total number of regions.

%Thirdly: the \emph{demes touching multiple regions ratio}.
%A deme may have generated individuals which will touch multiple regions.
%A high fraction of such demes suggests, that the algorithm doesn't separate
%basins of attraction well.
%It is a fraction of all demes satisfying this condition.

\begin{figure}
    \centering
    \begin{tikzpicture}
        \begin{groupplot}[group style={
                group size=3 by 1,
                horizontal sep=1.2cm,
                vertical sep=2cm},
            width=5.5cm,
            xlabel={budget},
            %view={0}{90},
            xmin=1000,xmax=17000,
            ymin=0,
            error bars/y dir=both, error bars/y explicit,
            legend entries={{NEA2}, {HMS}},
            legend pos=north east
            ]

            \nextgroupplot[
                title={Clusters after global phase}
            ]
            \addplot[only marks,mark=x] table [header=false, x index=1, y index=15, y error index=16] {data/benchmark4-case-II/log-2/metrics.nea2.data};
            \addplot[only marks,mark=o] table [header=false, x index=1, y index=15, y error index=16] {data/benchmark4-case-II/log-2/metrics.hms.data};

            \nextgroupplot[
                title={Clusters after reduction phase}
            ]
            \addplot[only marks,mark=x] table [header=false, x index=1, y index=19, y error index=20] {data/benchmark4-case-II/log-2/metrics.nea2.data};
            \addplot[only marks,mark=o] table [header=false, x index=1, y index=19, y error index=20] {data/benchmark4-case-II/log-2/metrics.hms.data};

            \nextgroupplot[
                title={Regions covered ratio},
                ymax=1
            ]
            \addplot[only marks,mark=x] table [header=false, x index=1, y index=23, y error index=24] {data/benchmark4-case-II/log-2/metrics.nea2.data};
            \addplot[only marks,mark=o] table [header=false, x index=1, y index=23, y error index=24] {data/benchmark4-case-II/log-2/metrics.hms.data};

            %\nextgroupplot[
                %title={Regions multiply touched ratio (lower better)},
                %ymin=0,ymax=1
            %]
            %\addplot[only marks,mark=x] table [header=false, x index=1, y index=27] {data/benchmark4-case-II/log-2/metrics.nea2.data};
            %\addplot[only marks,mark=o] table [header=false, x index=1, y index=27] {data/benchmark4-case-II/log-2/metrics.hms.data};

            %\nextgroupplot[
                %title={Demes touching multiple regions ratio (lower better)},
                %ymin=0,ymax=1
            %]
            %\addplot[only marks,mark=x] table [header=false, x index=1, y index=31] {data/benchmark4-case-II/log-2/metrics.nea2.data};
            %\addplot[only marks,mark=o] table [header=false, x index=1, y index=31] {data/benchmark4-case-II/log-2/metrics.hms.data};
        \end{groupplot}
    \end{tikzpicture}
    \caption{
      Metric values for NEA2 and HMS-CMA-ES variants in the second case.
      The data points are shown for budgets ranging from 2000 to 10000 evaluations.
      Each point was obtained by running the target configuration 10 times.
    }
    \label{fig:case-II-results}
\end{figure}
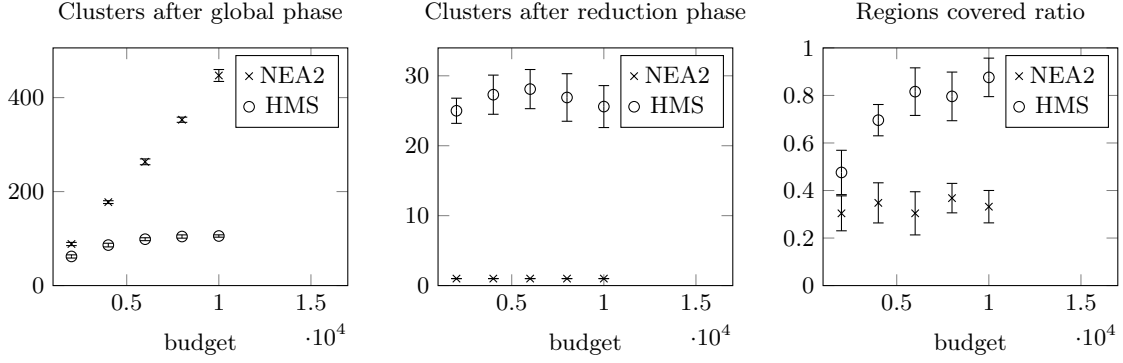

The graphs of the number of clusters after the global phase and after
reduction, and the ratio of covered regions are shown in
Figure~\ref{fig:case-II-results}.
The two first plots show how the algorithms behaved directly after the global
phase.
In global phase, HMS was much more conservative as the number of clusters go
--- which is understandable because of the limitation on sprout distances.
This is valuable because of the ability to cover a larger region with a similar
budget.

The issue with NEA2 clusters is that they don't separate the basins of
attraction well enough.
This leads the hill-valley reduction method to join all the clusters into a
single one.
NEA2 was designed with a focus on identification of localized minimizers in
mind and not as a global phase of insensitivity regions approximation strategy.
However, its interests are close enough to our approach to make sense to
compare it to our solution.

The regions covered ratio is shown in the third graph.
The results of HMS-CMA-ES variant increase with budget, while NEA2 variant
stagnates.

%The values of the three metrics described above favor HMS.
%The only metric in which NEA2 can be compared to HMS is the region covered
%ratio.
%It is visible there that with increasing budget HMS gets better while NEA2
%stays at the same ratio.

%As far as the other two metrics go, HMS results in quite a few multiply touched
%regions.
%This is most likely caused by the MWEA run, which is tasked at covering the
%neighborhood of insensitive regions, so it also happens to transfer several
%individuals to other basins of attraction.
%However, only about 30\% of demes touch more than one region, so most of the
%demes are correctly separated, with a few exceptions per run.
%But with each exception a number of regions is touched.

\begin{figure}
    \centering
    \begin{subfigure}[b]{0.49\textwidth}
        \includegraphics[width=\textwidth]{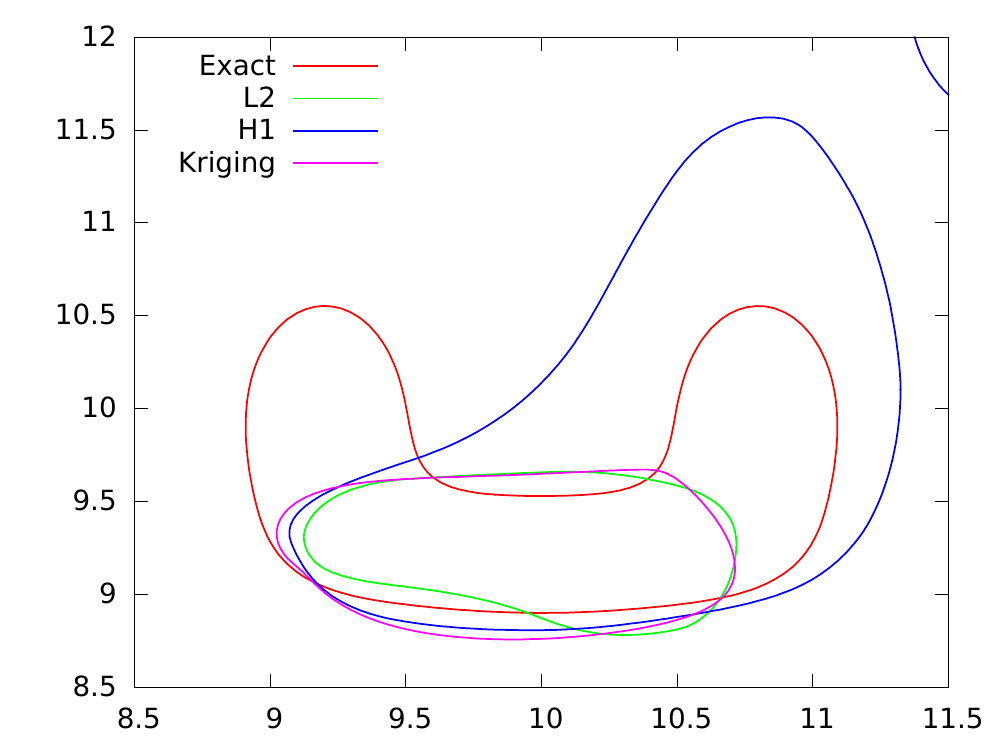}
        \caption{Resulting contours for NEA2}
    \end{subfigure}
    \begin{subfigure}[b]{0.49\textwidth}
        \includegraphics[width=\textwidth]{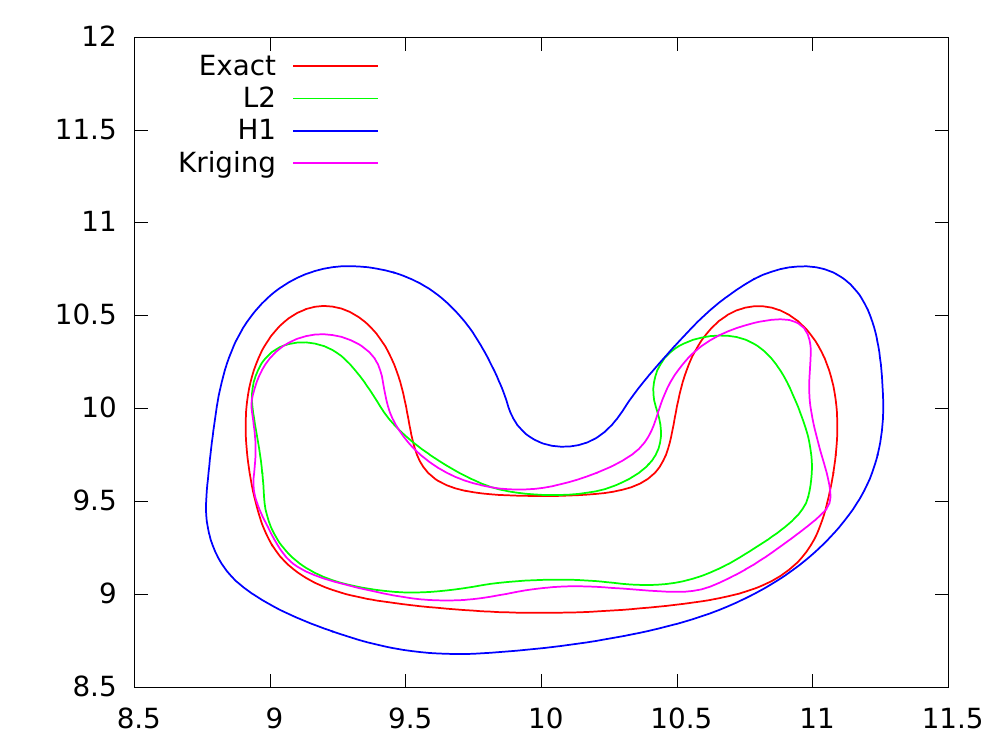}
        \caption{Resulting contours for HMS-CMA-ES}
    \end{subfigure}
    \caption{
        Each graph presents contours of the insensitivity region generated by
        approximation methods compared to the exact contour.
        $L^2$, $H^1$ and Kriging methods are shown.
    }
    \label{fig:case-II-results-approx}
\end{figure}

The level sets for the points from MWEA demes are shown in
Figure~\ref{fig:case-II-results-approx}.
Since there are many disjoint insensitivity regions with the same shape,
approximation of only one of these is presented.
Approximation based on points generated by HMS-CMA-ES is significantly better than in case of NEA2.

This is confirmed by the insensitivity region approximation quality data presented
in Table~\ref{tab:case-II-hausdorff}.
For each run and each of the 25 regions we computed Hausdorff distance between the exact insensitivity region
and its approximations, except for cases where the method failed to discover a region.
The table presents average distances for NEA2 and HMS-CMA-ES with different budgets and for various
approximation strategies.
While for $H1$-projections the differences are minor, when using~$L2$-projections,
and especially the Kriging method, HMS-CMA-ES approach is clearly superior.

\begin{table}
  \setlength{\tabcolsep}{5pt}
  \renewcommand{\arraystretch}{1.3}

  \centering
  \begin{tabular}{clrrr}
    \toprule
    \bf Algorithm & \bf Budget & \bf $L^2$-projection & \bf $H^1$-projection & \bf Kriging \\
    \midrule
    \multirow{5}{*}{NEA2}
        &  2,000 & $1.372 \pm 0.13$ & $1.910 \pm 0.35$ & $1.313 \pm 0.07$ \\
        &  4,000 & $1.369 \pm 0.12$ & $1.766 \pm 0.18$ & $1.315 \pm 0.11$ \\
        &  6,000 & $1.390 \pm 0.11$ & $1.707 \pm 0.18$ & $1.315 \pm 0.10$ \\
        &  8,000 & $1.395 \pm 0.08$ & $1.797 \pm 0.18$ & $1.309 \pm 0.06$ \\
        & 10,000 & $1.319 \pm 0.09$ & $1.746 \pm 0.21$ & $1.322 \pm 0.07$ \\
    \hline
    \multirow{5}{*}{HMS-CMA-ES}
        &  2,000 & $1.382 \pm 0.16$ & $1.761 \pm 0.19$ & $1.056 \pm 0.11$ \\
        &  4,000 & $1.062 \pm 0.09$ & $1.746 \pm 0.15$ & $0.797 \pm 0.05$ \\
        &  6,000 & $0.939 \pm 0.14$ & $1.781 \pm 0.11$ & $0.664 \pm 0.07$ \\
        &  8,000 & $0.907 \pm 0.13$ & $1.700 \pm 0.16$ & $0.610 \pm 0.08$ \\
        & 10,000 & $0.940 \pm 0.15$ & $1.770 \pm 0.20$ & $0.640 \pm 0.09$ \\
    \bottomrule

  \end{tabular}
  \\[6pt]
  \caption{Average Hausdorff distances between the exact sets of insensitivity and their approximations.}
  \label{tab:case-II-hausdorff}
\end{table}

\subsection{Third case - 4D Rastrigin}
\label{sec:case-III}

The fitness function in this case is defined as follows:
\begin{equation}
    \label{eq:f3}
    f_3\colon [-5,5]\times[-2,2]^3\ni (x_1,\dotsc,x_4) \to 2-\frac{1}{2}\left(\cos{\frac{\pi x_i}{5}}+\sum_{i=2}^4{\cos{\pi x_i}}\right) \in [0,4]\text{ .}
\end{equation}
The minima are placed on a regular grid with step size 10 in the first dimension and
distance 2 in the remaining ones.
There are 27 global minima in this case.

The settings of the algorithms remain the same as in the previous tests, apart
from the budget of the global phase, which is set to 50000 evaluations.
\rev{At such budget the performance of the methods approximately stagnates.}
We ran each algorithm 10 times to gather statistics.

The ratio of covered minima in this case is $44.1\%\pm0.6\%$ for NEA2 and $59\%\pm4\%$ for HMS-CMA-ES.
A minimum is considered to be covered by a deme if the deme has
generated an individual closer than 0.4 from the minimum.
%The ratio of minima covered more than once is equal to 4.0\% for NEA2 and 0 for HMS.
%The ratio of demes covering more than one minimum is 0 for NEA2 and 2.0\% for
%HMS.

Insensitivity region approximation results are presented in Table~\ref{tab:case-III-hausdorff}.
As in the previous benchmarks, for each of the 27 insensitivity regions we compare the exact region with
its approximation constructed as a level set of objective function approximation based on points generated
by NEA2 and HMS-CMA-ES.
For both algorithms the best results are obtained using Kriging method and in this case once again,
HMS-CMA-ES proves clearly superior.
It is worth to note that $H^1$-projection method performs relatively worse compared to other approximation
strategies, the difference being significantly more pronounced than in the previous benchmark.

\begin{table}
  \setlength{\tabcolsep}{5pt}
  \renewcommand{\arraystretch}{1.3}

  \centering
  \begin{tabular}{crrr}
    \toprule
    \bf Algorithm & \bf $L^2$-projection & \bf $H^1$-projection & \bf Kriging \\
    \midrule
      NEA2       & $1.426 \pm 0.08$ & $1.722 \pm 0.62$ & $1.406 \pm 0.09$ \\
      HMS-CMA-ES & $1.181 \pm 0.01$ & $2.618 \pm 0.68$ & $0.722 \pm 0.05$ \\
    \bottomrule

  \end{tabular}
  \\[6pt]
  \caption{Average Hausdorff distances between the exact sets of insensitivity and their approximations
           for 4D Rastrigin benchmark.}
  \label{tab:case-III-hausdorff}
\end{table}

\section{Conclusions and remarks}
\label{sec:Conc}

%We have presented an improvement of a strategy of determining shapes of
%insensitivity areas which surround global minimizers~\cite{Applanforms2018}.
%%
We have presented a new strategy of determining shapes of
insensitivity areas which surround global minimizers,
being a subsequent stage of our research~\cite{Applanforms2018}.
%%
%As the strategy is aimed at solving inverse problems which are inherently
%computationally-expensive to solve, it is desired to limit the number of
%necessary objective function evaluations in the process.
The first phase %stage 
of the strategy is global search, which aim is to identify
basins of attraction, a process which should make as few objective evaluations as
possible while maintaining the global search capabilities.
HMS invented by the authors was applied in this phase.
We have also used NEA2, which is well known to efficiently identify multiple
minima, as a reference algorithm.
The random sample gathered after the global phase is then transformed 
in local phase to better fill the regions of insensitivity and separate
their connected components.

Instead of using SEA in leaves of HMS, we substituted it by CMA-ES.
We exploit probability density information gathered during its convergence to
better localize the area of each separate, connected insensitivity region.
It allows us to abandon a more complex to configure density-clustering method.

We demonstrate the performance of our strategy on three benchmarks.
The first one shows the random sample transformation in its consecutive stages.
The second one allows to compare HMS-CMA-ES results with baseline, i.e. NEA2 for an
objective function with 25 separate regions of insensitivity.
The last one shows the efficiency of the strategy by solving a 4D problem with
27 minimizers.
Generally, the strategy using HMS-CMA-ES fills and separates the connected components
of insensitivity regions slightly better than the NEA2 one
(see Figures \ref{fig:case-I-results}, \ref{fig:case-II-results},
\ref{fig:case-II-results-approx}, Table \ref{tab:case-II-hausdorff},
\ref{tab:case-III-hausdorff}).

%We have found the improved strategy to fare better than the baseline strategy.
%Not only does it deliver better insensitivity region approximations, it also
%results in evaluation count reduced by a factor of 7.

%The benchmark we have presented is unimodal in terms of the number of
%insensitivity regions, however, we have tested the improved global phase as an
%optimization strategy and it was able to outperform the SEA variant on
%multimodal cases too (see \cite{CMA-ES-JOCS}).
%The results have been submitted as a full length paper to JoCS.

%Summing up, we have found that using CMA-ES as the final level of HMS strategy
%results in more efficient and more effective determination of the insensitivity
%regions surrounding global minimizers.

We plan to further improve the strategy to be able to handle multi-objective
problems.
It can be achieved by using MOEA algorithm with NSGA selection (see e.g.
\cite{Deb2002NSGA-II}), or a modified selection based on domination ranks (see
\cite{MOEA2017}). %,Gajda_PhD2015,MOEA_Proc_2015}).
The sets to be found in this case are associated with the connected components
of a Pareto set.

\bibliography{hms_spec_mo}

\end{document}